\documentclass[letterpaper]{article}
\usepackage[nopatch=hyphens]{microtype}
\usepackage{aaai2026}
\nocopyright
\usepackage{times}
\usepackage{helvet}
\usepackage{courier}
\usepackage[hyphens]{url}
\usepackage{graphicx}
\usepackage{natbib}
\usepackage{caption}
\usepackage{algorithm}
\usepackage{algorithmic}
\usepackage{amsmath}
\usepackage{amssymb}
\usepackage{booktabs}
\usepackage{multirow}
\usepackage{float}
\usepackage{placeins}
\usepackage{listings}
\usepackage{xcolor}

\definecolor{codegreen}{rgb}{0.15,0.55,0.15}
\definecolor{codegray}{rgb}{0.5,0.5,0.5}
\definecolor{codepurple}{rgb}{0.58,0,0.82}
\definecolor{backcolour}{rgb}{0.97,0.97,0.97}
\lstdefinestyle{python}{
    backgroundcolor=\color{backcolour},
    commentstyle=\color{codegreen},
    keywordstyle=\color{blue},
    stringstyle=\color{codepurple},
    basicstyle=\footnotesize\ttfamily,
    breakatwhitespace=false,
    breaklines=true,
    keepspaces=true,
    numbers=none,
    showspaces=false,
    showstringspaces=false,
    showtabs=false,
    tabsize=4,
    language=Python,
    frame=single,
    framerule=0.4pt,
    rulecolor=\color{codegray},
    aboveskip=6pt,
    belowskip=6pt,
    xleftmargin=3pt,
    xrightmargin=3pt,
}

\title{Beyond Static Evaluation: Co-Evolutionary Mechanisms \\
for LLM-Driven Strategy Evolution in Adversarial Games}

\author{
Haoran Li$^{1}$, Zengle Ge$^{2}$, Ziyang Zhang$^{2}$, Xiaomin Yuan$^{2}$, Yui Lo$^{3}$, Qianhui Liu$^{4}$, Bocheng An$^{5}$, Dongke Rong$^{2}$, Jiaqun Liu$^{6}$, Annan Li$^{2}$, Jianmin Wu$^{2}$, Dawei Yin$^{2}$, Dou Shen$^{2,\dagger}$
}

\affiliations{
$^1$ University of Chinese Academy of Sciences \\
$^2$ Famou Agent Team, Baidu AI Cloud \\
$^3$ The University of Sydney, Australia \\
$^4$ School of Artificial Intelligence, University of Chinese Academy of Sciences \\
$^5$ School of Transportation, Southeast University \\
$^6$ School of Software and Microelectronics, Peking University 
}

\begin{document}
\raggedbottom

\maketitle

\begin{abstract}
Recent advances in LLM-driven code evolution have enabled automated discovery by iteratively generating and improving programs.
However, applying these methods to adversarial multi-agent games introduces a fundamental challenge: the evaluation landscape shifts as strategies improve, causing fixed evaluators to become unreliable and evolution to stagnate.
We propose three mechanisms to address this challenge: \textbf{evaluator co-evolution}, which incorporates discovered champions into the opponent pool; \textbf{hierarchical deep evaluation}, which replaces noisy few-game scores with statistically reliable assessments; and \textbf{weakness pressure}, which dynamically up-weights the most difficult opponents to break through plateaus.

We implement these mechanisms within FAMOU, a framework built upon the same foundation-model code-evolution paradigm as OpenEvolve and ShinkaEvolve. On the MCTF 2026 3v3 maritime capture-the-flag task, FAMOU consistently outperforms both baselines under two backbone LLMs, achieving the highest combined score (0.526) and the best generalization to unseen opponents (61.7\% win rate), while ablations confirm that each mechanism contributes to performance.
Notably, the LLM mutation process generates tactical structures entirely absent from the seed strategies---including lookahead search and adaptive interception---demonstrating that code-level evolution can produce nontrivial algorithmic innovations in adversarial settings.
The FAMOU-evolved strategy further achieved 1st place in the hardware round-robin and 3rd in simulation at the AAMAS 2026 MCTF Competition, validating its real-world transferability. The optimized implementation and corresponding evaluation codes developed through our evolutionary process are available at: \url{https://github.com/1xiangliu1/FAMOU-CoEvo}
\end{abstract}

\section{Introduction}

\subsection{Background and Motivation}
Adversarial multi-agent games remain central in AI research \cite{busoniu2008comprehensive,hernandezleal2017survey}, combining non-stationary opponents, exponentially growing joint action spaces, and non-transitive dominance relations \cite{czarnecki2020real}.
These challenges are especially pronounced in team-based games where agents must coordinate within their team while competing against adversaries.

We study these challenges through MCTF 2026 (Maritime Capture The Flag), a 3v3 maritime capture-the-flag competition on a 160\,m $\times$ 80\,m field where teams are ranked by total captures.
In our early competition attempts, the reinforcement learning policies we trained failed to outperform handwritten heuristic strategies.
This motivated a different approach: instead of learning policies from scratch, we use LLM-driven code-level evolution to automatically improve the heuristic code itself, preserving its structural advantages while pushing performance beyond what manual design can achieve.

\subsection{Code-Level Evolution Approach}
Inspired by FunSearch \cite{romera2024funsearch}, ELM \cite{lehman2022elm}, and Famou-Agent \cite{li2026famou}, we search directly over complete strategy code (500--1700 lines) by using LLM-generated semantic mutations and evaluator-driven selection with co-evolutionary dynamics. Unlike end-to-end reinforcement learning, this approach preserves the interpretable structure of heuristic code while allowing the LLM to introduce new tactical logic absent from the seeds---such as lookahead search or dynamic role auctions.

\subsection{Contributions}
This paper makes the following contributions:
\begin{enumerate}
\item \textbf{A systematic framework comparison for LLM code-level evolution in adversarial games.}
We compare FAMOU with OpenEvolve \cite{sharma2025openevolve} and ShinkaEvolve \cite{lange2025shinkaevolve} across two backbone LLMs using standardized MCTF evaluation. FAMOU consistently outperforms both baselines.

\item \textbf{Evaluator co-evolution with deep evaluation and weakness pressure.}
We introduce three mechanisms---evaluator co-evolution (incorporating discovered champions into the opponent pool), hierarchical deep evaluation (replacing noisy few-game scores with statistically reliable assessments), and weakness pressure (dynamically up-weighting the most difficult opponents)---and assess their contributions through exploratory ablations.

\item \textbf{Empirical evidence of LLM-generated tactical structures.}
We document LLM-generated tactical structures absent from the seed code, including H-DWA (lookahead search), A-Lock (role locking), and K-Filter (EWMA-based interception), providing evidence that LLMs can generate nontrivial algorithmic structures in adversarial games.
\end{enumerate}

\section{Related Work}

\subsection{Multi-Agent Adversarial Games}
Research on multi-agent adversarial games spans heuristic and learning-based paradigms \cite{busoniu2008comprehensive,hernandezleal2017survey}.
Heuristic strategies (e.g., hierarchical state machines \cite{kalyanakrishnan2006robocup}, artificial potential fields \cite{khatib1986real}) offer natural advantages in determinism and interpretability.
Deep RL systems such as AlphaStar \cite{vinyals2019grandmaster}, OpenAI Five \cite{berner2019dota}, and DeepMind CTF \cite{jaderberg2019ctf}, together with multi-agent algorithms such as QMIX \cite{rashid2018qmix} and MAPPO \cite{yu2022mappo} and paradigms such as fictitious self-play \cite{heinrich2015fictitious}, PSRO \cite{lanctot2017unified}, and population-based training \cite{jaderberg2017population}, have achieved superhuman performance in complex games.
However, the inherent non-transitivity of adversarial games \cite{czarnecki2020real,balduzzi2019open} makes it difficult to find globally robust strategies.

\subsection{LLM-Driven Code Generation and Program Synthesis}
From Codex \cite{chen2021codex} to DeepSeek-Coder \cite{guo2024deepseek}, the code-generation capabilities of LLMs have advanced rapidly \cite{li2022alphacode,li2023starcoder,roziere2023codellama}.
Programming agents such as SWE-agent \cite{yang2024sweagent} and Voyager \cite{wang2023voyager} demonstrate that LLMs can perform multi-step software-engineering and open-world exploration tasks.
At the intersection of LLMs and evolutionary search, FunSearch \cite{romera2024funsearch} was the first to combine LLMs with evolutionary search and surpass human-best solutions in mathematical discovery.
ELM \cite{lehman2022elm} proposed LLMs as intelligent mutation operators, ReEvo \cite{ye2024reevo} introduced reflective mechanisms to enhance heuristic evolution, and Eureka \cite{ma2023eureka} used LLMs to automatically design RL reward functions.
Famou-Agent \cite{li2026famou} combines LLM-based evolution with evaluation-feedback loops in scientific computing, serving as the direct foundation for the framework in this paper.

At the intersection of LLM evolution and multi-agent games, PolicyEvolve \cite{lv2025policyevolve} was the first to propose a programmatic strategy evolution framework for multiplayer games, achieving continuous strategy improvement through global/local strategy pools and population-based training.
COvolve \cite{sygkounas2026covolve} models LLM-generated strategies and environments as a zero-sum game, improving strategy robustness through adversarial co-evolution.

\subsection{Evolutionary Search and Quality-Diversity}
Classical genetic programming (GP) \cite{koza1992genetic} has a long history in program synthesis tasks.
NEAT \cite{stanley2002neat} evolves neural network structures through topology augmentation, and regularized evolution \cite{real2019regularized} combines tournament selection with age regularization for architecture search.
The island model \cite{whitley1999island} effectively balances exploration and exploitation by maintaining multiple independent subpopulations with periodic elite migration.
MAP-Elites \cite{mouret2015mapelites} and NSGA-II \cite{deb2002nsga} further expand the dimensions of quality-diversity search.

\subsection{Co-Evolution and Adversarial Evaluation}
The central idea of co-evolution is that test cases and evaluated subjects evolve together, forming a continuously escalating ``arms race'' \cite{hillis1990coevolving}.
Rosin and Belew \cite{rosin1997new} systematically studied evaluation difficulties in competitive co-evolution---including the ``Red Queen effect'' and cyclic dominance---and proposed mitigation methods such as competitive fitness sharing.
Rating systems like Elo \cite{elo1978rating} and TrueSkill \cite{herbrich2006trueskill} attempt to infer true strategy strength from limited match data, but the non-transitivity of adversarial games \cite{czarnecki2020real} makes evaluation based on limited opponent pools inherently unreliable.
The evaluator co-evolution proposed in this paper can be viewed as an instantiation of competitive co-evolution in the domain of LLM code evolution.

\textbf{Positioning.}
Our work differs from prior LLM-based evolution in three respects:
(1) the evolution target is a complete 500--1700-line strategy system rather than a compact function;
(2) we introduce evaluator co-evolution, deep evaluation, and weakness pressure to combat opponent-pool staleness;
(3) we provide controlled experiments with statistical significance testing across two backbone LLMs.

\section{Problem Definition and Challenges}

\subsection{MCTF 2026 Competition Rules}
MCTF 2026 is a 3v3 maritime capture-the-flag competition. The rules are summarized in Table~\ref{tab:rules}.

\begin{table}[!ht]
\centering
\caption{Summary of the MCTF 2026 competition rules.}
\label{tab:rules}
\small
\begin{tabular}{@{}lp{5.5cm}@{}}
\toprule
\textbf{Attribute} & \textbf{Description} \\
\midrule
Field & 160\,m $\times$ 80\,m, midline $x{=}80$ \\
Teams & 3v3 agents \\
Flags & Blue $(0,40)$, Red $(160,40)$ \\
Grab & Enter 10\,m of enemy flag ($+0.1$) \\
Score & Carry flag to own corner $(0,0)$ or $(0,80)$ within 20\,m ($+1.0$) \\
Tag & Tag enemy within 10\,m in own territory, causing respawn (60\,s cooldown) \\
PowerPlay & Random periods disabling one enemy \\
Duration & 600 seconds per game \\
Ranking & Total captures (not win rate) \\
Actions & $\{0,\ldots,23\}$: 0--22 headings (speed 1.0), 23 stop \\
\bottomrule
\end{tabular}
\end{table}

Participants submit Python strategy code for the blue team, which the platform evaluates against all opponents.
Strategies are ranked by total captures, meaning that a 5--0 victory is far more valuable than a narrow 1--0 win.

\subsection{Core Challenges in Strategy Design}
MCTF strategy design faces five core challenges:
(1) \emph{high-dimensional action space}: $24^3 = 13{,}824$ joint action combinations;
(2) \emph{role assignment}: three agents must be dynamically assigned to roles such as attacker, defender, and support;
(3) \emph{offense-defense balance}: ranking by total captures requires maximizing scoring efficiency while minimizing concessions;
(4) \emph{opponent adaptability}: opponent strategies are unknown and diverse, requiring good generalization;
(5) \emph{non-transitivity}: no globally optimal strategy exists.

\section{Method}

\subsection{Overall Framework}
\begin{figure*}[t]
\centering
\includegraphics[width=0.85\textwidth]{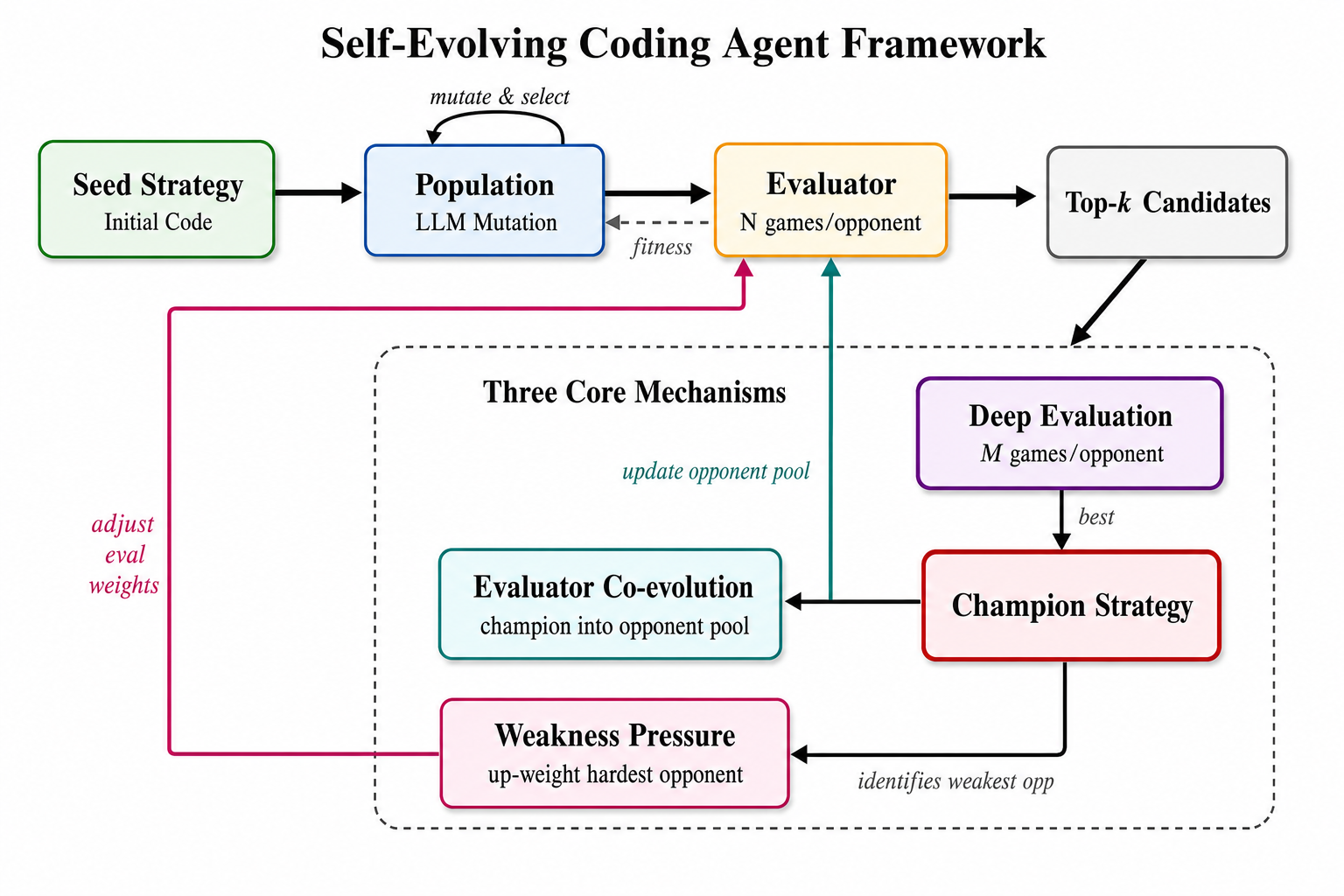}
\caption{FAMOU self-evolving coding-agent framework. Seed strategies undergo LLM-based semantic mutation and evolution; an evaluator screens candidates; hierarchical deep evaluation selects champions; champions are automatically added to the next evaluator's opponent pool through co-evolution; and weakness pressure dynamically adjusts opponent weights to overcome plateaus.}
\label{fig:framework}
\end{figure*}

We use FAMOU (Framework for Automated Mutation and Optimization of Utilities) \cite{li2026famou} to optimize executable MCTF strategy code with LLM-generated semantic mutations.
Relative to Famou-Agent \cite{li2026famou}, we adapt the framework to adversarial multi-agent strategy code and add evaluator co-evolution, weakness pressure, and MCTF experiments.
The general LLM mutation loop is inherited from Famou-Agent: the system maintains an archive of executable strategy programs, selects high-scoring programs as parents, prompts the LLM to produce complete modified Python files rather than patches, validates the generated files for syntax and API compatibility, evaluates valid candidates in the game simulator, and feeds the resulting performance summary back into the next mutation prompt.
Figure~\ref{fig:framework} shows the architecture, and Algorithm~\ref{alg:main} gives the core process.
The remainder of this section details the key components: the evolution target and mutation operator (\S4.2), parent selection (\S4.3), and the three mechanisms that distinguish FAMOU from vanilla LLM evolution---evaluator co-evolution (\S4.4), weakness pressure (\S4.5), and hierarchical deep evaluation (\S4.6).

\begin{algorithm*}[t]
\caption{Main Loop for MCTF Strategy Evolution}
\label{alg:main}
\textbf{Require:} Seed strategy $S_0$, evaluator $\mathcal{E}$, number of iterations $T$ \quad
\textbf{Ensure:} Best strategy $S^*$
\begin{algorithmic}[1]
\STATE Initialize population $P_0$ containing $S_0$ variants
\FOR{$t = 1$ to $T$}
  \FOR{each candidate $c \in P_t$}
    \STATE Run $c$ against each opponent in $\mathcal{E}$ (3 games/opponent); compute weighted fitness $F(c) = \sum_{i=1}^{n} w_i \cdot \text{metric}(c, o_i)$
  \ENDFOR
  \STATE Select elite individuals based on $F(c)$; LLM performs semantic mutation on elite code $\to$ new candidates
  \IF{champion detected (new best)}
    \STATE Add champion to the evaluator opponent pool (\textbf{co-evolution}); apply \textbf{weakness pressure}: identify and up-weight weakest opponent
    \STATE Trigger \textbf{deep evaluation} (20 games/opponent) for confirmation
  \ENDIF
\ENDFOR
\STATE \textbf{return} Best strategy $S^*$ from deep evaluation
\end{algorithmic}
\end{algorithm*}

\subsection{Evolution Target and LLM Mutation Operator}
Each individual is a 500--1700-line Python strategy file with an \texttt{Agent\_0.compute\_action(obs, info)} method.
Mutations are performed by LLMs, specifically Gemini-2.5-Flash or DeepSeek-V4-Flash, with temperature 0.8 and a maximum token budget of 64,000.
For each mutation, the LLM receives the current strategy, task/API constraints, and per-opponent evaluation feedback, then returns a complete executable strategy file preserving the required interface.

LLM mutation uses code semantics to make directed structural changes rather than random perturbations.
Evolution starts from four handwritten heuristic seed strategies (Table~\ref{tab:seeds}), covering different tactical styles including potential field avoidance, auction-based assignment, and lane-based offense.

\begin{table}[!ht]
\centering
\caption{Handwritten heuristic seed strategies used to initialize evolution.}
\label{tab:seeds}
\small
\begin{tabular}{@{}lcp{4.0cm}@{}}
\toprule
\textbf{Seed} & \textbf{vs.\ hard} & \textbf{Core mechanism} \\
\midrule
Balanced & ${\sim}100\%$ & Role assignment + potential field + curved interception \\
Agile & ${\sim}95\%$  & Distributed auction + vortex avoidance \\
Structured & ${\sim}90\%$  & Three-lane offense + zone defense \\
Safe & ${\sim}85\%$  & Beam search return + momentum integration \\
\bottomrule
\end{tabular}
\end{table}

\subsection{Parent Selection via UCB}
To balance the exploitation of high-fitness programs with the exploration of under-tested candidates, we select parent programs for mutation using an Upper Confidence Bound (UCB) bandit policy \cite{auer2002finite}:
\begin{equation}
\text{UCB}(p_i) \;=\; \bar{s}_i \;+\; c \,\sqrt{\frac{2\,\ln N}{n_i}}
\label{eq:ucb}
\end{equation}
where $\bar{s}_i$ is the mean fitness score of program $p_i$, $N$ is the total number of parent selections so far, $n_i$ is the number of times $p_i$ has been selected as a parent, and $c$ is the exploration coefficient.
The first term favors high-performing programs, while the second term provides an exploration bonus that grows with under-sampling, ensuring that promising but less-explored programs are revisited.

\subsection{Evaluator Co-Evolution}
Once a parent is selected and mutated, the resulting candidate must be evaluated. In adversarial games, a fixed evaluator can become stale as strategies improve, so we treat the evaluator and strategies as co-evolving populations \cite{hillis1990coevolving,rosin1997new}. We define the evaluator as follows:

\textbf{Definition 1} (Evaluator). An \emph{evaluator} $\mathcal{E} = (O, \mathbf{w}, F)$ consists of an opponent pool $O = \{o_1, \ldots, o_n\}$, weights $\mathbf{w} = [w_1, \ldots, w_n]$ ($\sum w_i = 1$), and a fitness function $F$:
\begin{equation}
F(c) = \sum_{i=1}^{n} w_i \cdot \text{metric}(c, o_i)
\end{equation}
where \emph{metric} denotes either win rate or scoring efficiency $\min(\text{avg\_score}/5.0, 1.0)$.

Whenever a new champion $C_t$ is produced, it becomes a high-weight ``gatekeeper'' in $\mathcal{E}_{t+1}$, so later candidates must outperform both original opponents and all previous champions.

\subsection{Weakness Pressure}
Co-evolution enriches the opponent pool, but does not address which opponents matter most. When a plateau reflects a persistent weak opponent, we apply weakness pressure to redirect selective pressure:
\begin{enumerate}
\item Perform a comprehensive deep evaluation of the current champion to identify the opponent $o_{\text{weak}}$ with the lowest win rate.
\item Double $o_{\text{weak}}$'s weight and renormalize:
\begin{equation}
w_{\text{weak}} \leftarrow 2\,w_{\text{weak}}
\label{eq:weakness1}
\end{equation}
\begin{equation}
w_j \leftarrow w_j \cdot \frac{1 - 2\,w_{\text{weak}}}{1 - w_{\text{weak}}} \;\;\forall\, j \neq \text{weak}
\label{eq:weakness2}
\end{equation}
\end{enumerate}
This makes improvement against the weakest opponent the main path to higher fitness.

\subsection{Hierarchical Deep Evaluation}
Both co-evolution and weakness pressure depend on accurate fitness estimates. However, fast evaluation scores can be unreliable: in preliminary experiments, 3-game scores had Spearman $\rho = 0.11$, $p = 0.69$ with 20--40-game deep-evaluation scores. This motivates a hierarchical system:
\begin{itemize}
\item \textbf{Fast evaluation} (3 games/opponent): for coarse population ranking during evolution.
\item \textbf{Deep evaluation} (20 games/opponent): for true score confirmation after candidate extraction.
\end{itemize}
Final decisions use deep evaluation; 3-game scores provide only coarse population ranking.

\section{Experiments and Analysis}

We conduct a systematic empirical study addressing three research questions:
\begin{itemize}
\item \textbf{RQ1}: Does FAMOU with the proposed mechanisms (evaluator co-evolution, deep evaluation, weakness pressure) outperform other code-level evolution frameworks (OpenEvolve, ShinkaEvolve) for adversarial games?
\item \textbf{RQ2}: What is the contribution of each proposed mechanism (evaluator co-evolution, deep evaluation, weakness pressure)?
\item \textbf{RQ3}: How well do evolved strategies generalize to unseen opponents?
\end{itemize}

\subsection{Experimental Setup}

\subsubsection{Experiment Configurations.}
We run six three-run main experiments and four single-run ablations, each for 400 iterations from the four seed strategies in Table~\ref{tab:seeds}.
All three frameworks (FAMOU, OpenEvolve, ShinkaEvolve) are tested under two backbone LLMs---DeepSeek-V4-Flash and Gemini-2.5-Flash---with the same seed strategies, iteration budget, and final benchmark protocol.
The experiments are organized into four groups:
\begin{itemize}
\item \textbf{Full framework} (2 experiments $\times$ 3 runs): FAMOU with all three mechanisms.
\item \textbf{Baseline: OpenEvolve} (2 experiments $\times$ 3 runs): OpenEvolve \cite{sharma2025openevolve}, a state-of-the-art LLM code evolution framework using single-population evolution with a fixed evaluator.
\item \textbf{Baseline: ShinkaEvolve} (2 experiments $\times$ 3 runs): ShinkaEvolve \cite{lange2025shinkaevolve}, an LLM code evolution framework emphasizing sample efficiency and open-ended search, employing island-model populations, novelty rejection sampling, and adaptive LLM ensemble selection.
\item \textbf{Ablation} (4 experiments, single run): FAMOU under Gemini-2.5-Flash with one mechanism removed at a time.
  \begin{itemize}
  \item w/o Co-evolution: remove evaluator co-evolution (fixed opponent pool)
  \item w/o Deep Eval: remove hierarchical deep evaluation (use only 3-game scores)
  \item w/o Weakness: remove weakness pressure (uniform opponent weights)
  \item Vanilla: remove all three mechanisms (basic LLM evolution with fixed evaluator)
  \end{itemize}
\end{itemize}

The key difference is that neither OpenEvolve nor ShinkaEvolve adds evolved champions to the evaluator opponent pool, applies weakness-pressure reweighting, or uses hierarchical deep evaluation during evolution.

\subsubsection{Benchmark Evaluation Protocol.}
The main comparison is equal-iteration ($T=400$), not equal-compute, because FAMOU adds deep-evaluation and co-evolution overhead. At every 40-iteration checkpoint, all experiments use the same benchmark:
\begin{itemize}
\item \textbf{10 benchmark opponents}: five seen (hard, Balanced, Agile, Structured, Safe) plus five held-out unseen opponents used only for checkpoint logging and final evaluation.
\item \textbf{20 games per opponent}: sufficient to reduce noise while remaining computationally feasible.
\end{itemize}

The \textbf{Combined Score (CS)} aggregates win rate and scoring margin:
\begin{equation}
\text{CS} = 0.7 \times \text{WR} + 0.3 \times \min\!\Big(1.0,\; \frac{\max(0,\;\text{margin})}{5.0}\Big)
\label{eq:cs}
\end{equation}
where WR denotes win rate and margin denotes the average score differential. The 0.7/0.3 weighting prioritizes reliable wins while rewarding high-margin scoring; the margin term is clipped at 5.0.

\subsubsection{Statistical Testing.}
We use the following statistical tests to assess significance:
\begin{itemize}
\item \textbf{Wilcoxon signed-rank test}: based on per-opponent win rate differences (paired, non-parametric).
\item \textbf{Paired $t$-test}: on per-opponent margin differences.
\item \textbf{Bootstrap 95\% confidence intervals} (10,000 resamples): for Combined Score.
\end{itemize}

\subsection{Main Results: FAMOU vs.\ Baselines (RQ1)}
Table~\ref{tab:main} reports final iteration-400 performance.

\begin{table}[t]
\centering
\caption{Final performance at iteration 400 (mean $\pm$ standard deviation over 3 runs, 10 benchmark opponents, 20 games per opponent). CS = Combined Score. The best result in each column is shown in \textbf{bold}.}
\label{tab:main}
\begingroup
\setlength{\tabcolsep}{3pt}%
\small
\resizebox{\columnwidth}{!}{%
\begin{tabular}{@{}llcccc@{}}
\toprule
\textbf{Method} & \textbf{LLM} & \textbf{CS} & \textbf{WR} & \textbf{Margin} & \textbf{Unseen WR} \\
\midrule
FAMOU & DeepSeek-V4-Flash & $\mathbf{0.526 \pm 0.075}$ & $\mathbf{0.680 \pm 0.080}$ & $\mathbf{+0.83 \pm 0.32}$ & $\mathbf{0.617 \pm 0.072}$ \\
FAMOU & Gemini-2.5-Flash & $0.505 \pm 0.108$ & $0.653 \pm 0.123$ & $+0.81 \pm 0.38$ & $0.530 \pm 0.144$ \\
\midrule
ShinkaEvolve & DeepSeek-V4-Flash & $0.437 \pm 0.028$ & $0.561 \pm 0.033$ & $+0.73 \pm 0.09$ & $0.342 \pm 0.054$ \\
ShinkaEvolve & Gemini-2.5-Flash & $0.319 \pm 0.139$ & $0.429 \pm 0.177$ & $+0.31 \pm 0.26$ & $0.335 \pm 0.110$ \\
\midrule
OpenEvolve & DeepSeek-V4-Flash & $0.382 \pm 0.046$ & $0.500 \pm 0.051$ & $+0.53 \pm 0.19$ & $0.312 \pm 0.041$ \\
OpenEvolve & Gemini-2.5-Flash & $0.257 \pm 0.025$ & $0.357 \pm 0.034$ & $+0.13 \pm 0.03$ & $0.262 \pm 0.052$ \\
\bottomrule
\end{tabular}%
}
\endgroup
\end{table}

The key findings are as follows:
\begin{itemize}
\item \textbf{FAMOU outperforms both baselines across both LLM backbones.}
Under both DeepSeek-V4-Flash and Gemini-2.5-Flash, FAMOU achieves higher CS, WR, and margin than ShinkaEvolve and OpenEvolve.
The best configuration—FAMOU under DeepSeek-V4-Flash—reaches CS $= 0.526$, a relative improvement of 20.4\% over the strongest baseline, ShinkaEvolve under the same backbone (CS $= 0.437$).

\item \textbf{LLM backbone matters.}
DeepSeek-V4-Flash consistently outperforms Gemini-2.5-Flash across all three frameworks, suggesting that stronger code-generation capability translates directly into better evolved strategies.

\item \textbf{FAMOU generalizes better to unseen opponents.}
Under DeepSeek-V4-Flash, FAMOU achieves the highest unseen win rate ($0.617$), while ShinkaEvolve drops sharply from $0.780$ seen to $0.342$ unseen, suggesting that co-evolutionary evaluation reduces overfitting to training opponents.

\item \textbf{FAMOU sustains productive search.}
Over 400 iterations, FAMOU under DeepSeek-V4-Flash discovers 10 unique champions with a 22.5\% stagnation rate, compared to only 3 champions and 65\% stagnation for OpenEvolve under Gemini-2.5-Flash, indicating that the co-evolutionary mechanisms maintain selective pressure against diminishing returns.
\end{itemize}

\subsection{Learning Curves}
\begin{figure}[t]
\centering
\includegraphics[width=0.95\columnwidth]{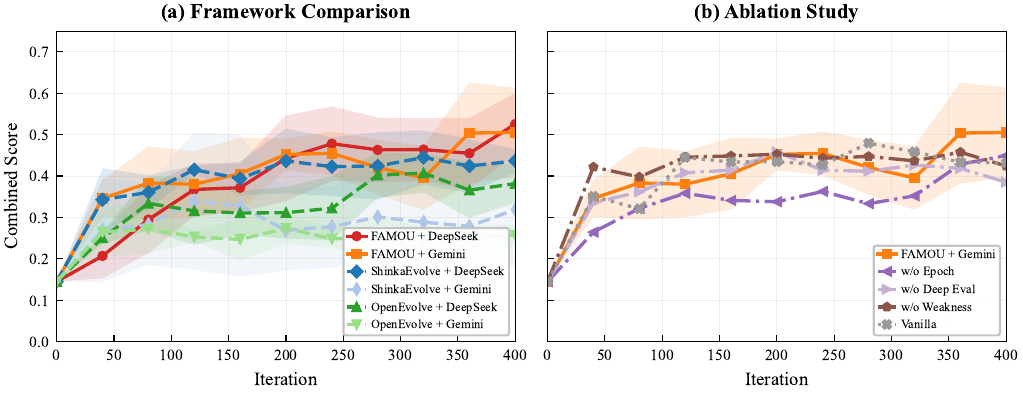}
\caption{Combined Score learning curves over 400 iterations, evaluated every 40 iterations on the 10-opponent benchmark with 20 games per opponent. Lines show the mean over 3 runs, and shaded regions show $\pm$ one standard deviation. FAMOU variants show sustained improvement, whereas the baselines plateau early.}
\label{fig:curves}
\end{figure}

As shown in Figure~\ref{fig:curves}, FAMOU variants continue improving through iteration 400, whereas baselines largely stagnate after iteration 240--280.

\subsubsection{Evolution Trajectory Analysis.}
\begin{figure}[t]
\centering
\includegraphics[width=0.95\columnwidth]{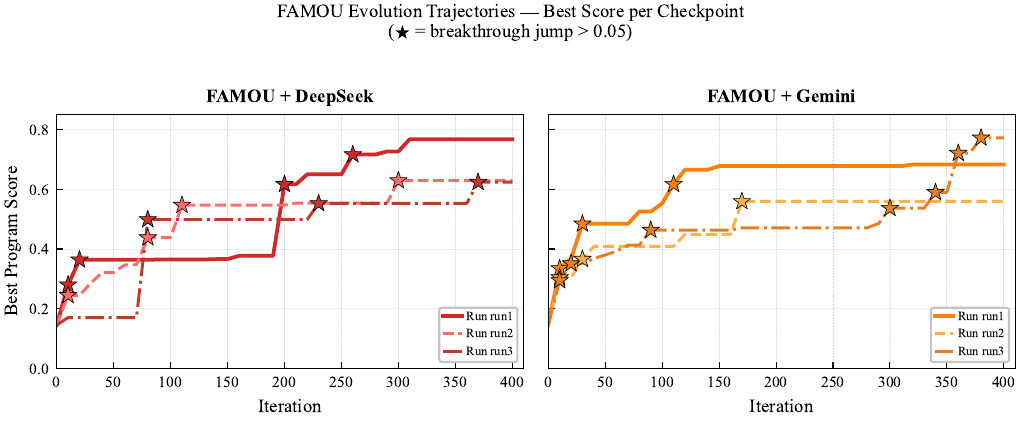}
\caption{FAMOU evolution trajectories based on the best score at 10-iteration resolution. All runs start from the same four-seed initialization; the best initial seed has CS $= 0.145$. Stars ($\star$) mark breakthroughs, defined as score increases greater than 0.05. Evolution exhibits punctuated equilibrium, with long plateaus interrupted by sudden jumps.}
\label{fig:evolution_detail}
\end{figure}

Figure~\ref{fig:evolution_detail} plots best-program scores every 10 iterations for all 6 FAMOU runs, revealing \emph{punctuated equilibrium}---long stagnation interrupted by sudden breakthroughs.
For example, under DeepSeek-V4-Flash, FAMOU Run 1 jumps from 0.377 to 0.617 at iteration 200, a $+63.7\%$ increase in a single step.
Under Gemini-2.5-Flash, FAMOU Run 3 shows three consecutive breakthroughs between iterations 340--380, increasing rapidly from 0.54 to 0.77.
The timing of these breakthroughs varies widely across runs (iterations 80 to 370), reflecting the inherent stochasticity of LLM-driven code generation.
Several breakthroughs occur after iteration 300, indicating that 400 iterations may be insufficient for full convergence.

\subsection{Ablation Study (RQ2)}
\begin{figure}[t]
\centering
\includegraphics[width=0.95\columnwidth]{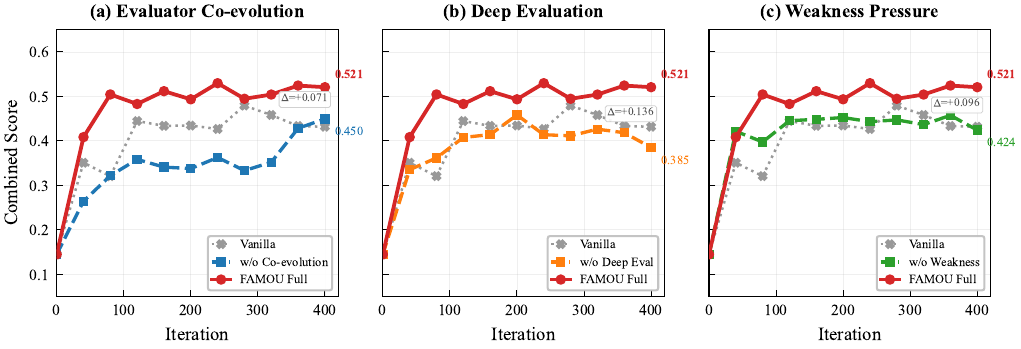}
\caption{Ablation learning curves (Gemini-2.5-Flash, single run). Each panel isolates one mechanism by comparing FAMOU Full (red) against the ablated variant (dashed) and Vanilla baseline (gray). Removing deep evaluation causes the largest degradation.}
\label{fig:ablation_curves}
\end{figure}

Table~\ref{tab:ablation} reports the ablation results, and Figure~\ref{fig:ablation_curves} shows the corresponding learning curves.

\begin{table}[t]
\centering
\caption{Exploratory ablation study (Gemini-2.5-Flash backbone, single run). $\Delta$CS is relative to the FAMOU Full configuration. $p$-values are from Wilcoxon signed-rank tests on per-opponent win rates. The best CS is shown in \textbf{bold}. Because each ablation was run once, results should be interpreted as suggestive rather than conclusive.}
\label{tab:ablation}
\small
\begin{tabular}{@{}lcccc@{}}
\toprule
\textbf{Configuration} & \textbf{CS} & $\boldsymbol{\Delta}$ \textbf{CS} & $\boldsymbol{\Delta}$\textbf{\%} & \textbf{$p$-value} \\
\midrule
FAMOU Full & $\mathbf{0.521}$ & --- & --- & --- \\
\midrule
w/o Co-evolution & 0.450 & $-0.071$ & $-13.7\%$ & 0.084 \\
w/o Deep Eval & 0.385 & $-0.136$ & $-26.0\%$ & 0.002 \\
w/o Weakness & 0.424 & $-0.097$ & $-18.6\%$ & 0.160 \\
Vanilla & 0.432 & $-0.089$ & $-17.1\%$ & 0.020 \\
\bottomrule
\end{tabular}
\end{table}

\begin{itemize}
\item \textbf{Removing deep evaluation produced the largest drop} ($\Delta\text{CS} = -0.136$, $p = 0.002$), consistent with the unreliability of 3-game selection.

\item \textbf{The full configuration outperforms the vanilla baseline} ($\Delta\text{CS} = +0.089$, $p = 0.020$), suggesting interaction among the three mechanisms.

\item \textbf{The mechanisms exhibit non-additive interactions.}
The sum of individual drops ($0.071 + 0.136 + 0.097 = 0.304$) exceeds the gap between the full configuration and the vanilla baseline (0.089), indicating that the effects of the mechanisms are not additive and that mechanisms may partially compensate for one another when combined.
\end{itemize}

\subsection{Cross-Evaluation Analysis}
\begin{figure}[htbp]
\centering
\includegraphics[width=0.95\columnwidth]{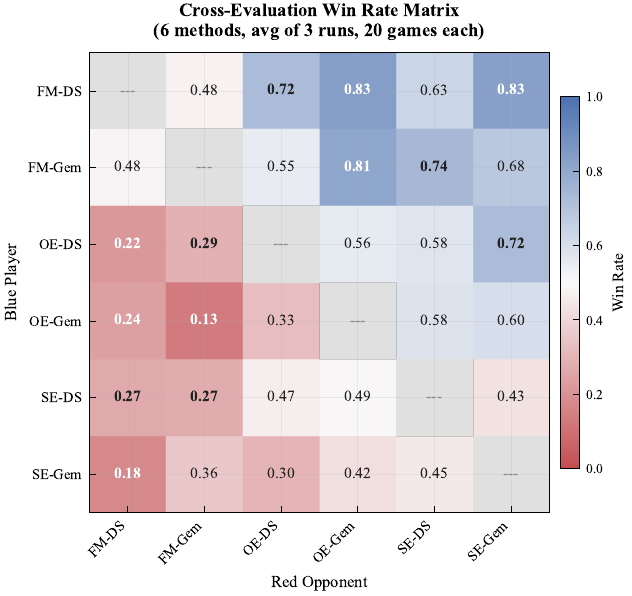}
\caption{Cross-evaluation win-rate matrix ($6 \times 6$, averaged over 3 runs with 20 games per directed matchup). Rows indicate the blue player, columns indicate the red opponent, and each cell reports the row strategy's win rate against the column strategy. FAMOU variants achieve the strongest average performance and consistently outperform the baseline variants.}
\label{fig:heatmap}
\end{figure}

To assess robustness beyond fixed-opponent benchmarks, we conduct a $6 \times 6$ round-robin cross-evaluation among the best strategies produced by each method (20 games per directed matchup, 1,800 total games across 3 runs; Figure~\ref{fig:heatmap}).
Under DeepSeek-V4-Flash, FAMOU achieves the highest average win rate (0.698), confirming the benchmark ranking.
Notably, the cross-evaluation reveals that benchmark scores do not fully capture pairwise relationships: OpenEvolve under DeepSeek-V4-Flash outperforms ShinkaEvolve in direct matchups despite lower benchmark scores, suggesting different specialization profiles that a diverse opponent pool can expose.

\subsection{Generalization Analysis (RQ3)}
\begin{figure}[htbp]
\centering
\includegraphics[width=0.95\columnwidth]{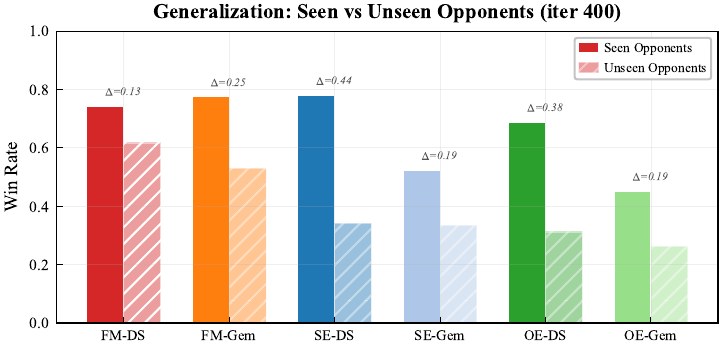}
\caption{Generalization analysis on the 10-opponent benchmark. Each point represents one method's mean win rate over 3 runs against the five seen training opponents (x-axis) and the five held-out unseen opponents (y-axis). Points below the diagonal indicate lower performance on unseen opponents than on seen opponents.}
\label{fig:generalization}
\end{figure}

Figure~\ref{fig:generalization} compares seen vs.\ unseen opponent performance:
\begin{itemize}
\item Under DeepSeek-V4-Flash, FAMOU achieves the best generalization with the smallest seen--unseen gap ($0.743 \pm 0.141$ vs.\ $0.617 \pm 0.072$), and lies closest to the diagonal.
\item Under the same backbone, ShinkaEvolve has high seen performance ($0.780$) but a large generalization gap (unseen $= 0.342$), indicating overfitting to training opponents.
\item OpenEvolve under Gemini-2.5-Flash shows relatively low performance on both seen and unseen opponents ($0.452$ vs.\ $0.262$)---it does not overfit because it never learns strong strategies.
\end{itemize}

\begin{figure}[htbp]
\centering
\includegraphics[width=0.95\columnwidth]{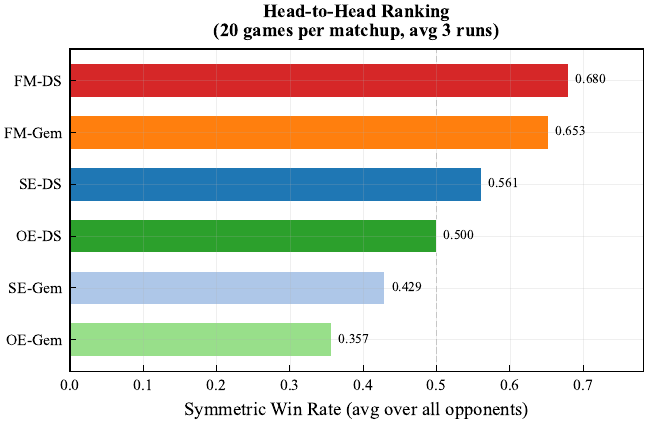}
\caption{Per-opponent win rates on the 10-opponent benchmark, averaged over 3 runs with 20 games per opponent. The left five are seen training opponents; the right five are held-out unseen opponents.}
\label{fig:ranking}
\end{figure}

The per-opponent breakdown in Figure~\ref{fig:ranking} further confirms this trend: under DeepSeek-V4-Flash, FAMOU achieves $>70\%$ win rate against eight of the ten opponents, while OpenEvolve under Gemini-2.5-Flash struggles against most unseen opponents (many $<30\%$).

\subsection{LLM-Invented Tactical Innovations}
During evolution, the LLM generated decision architectures absent from the seeds (Table~\ref{tab:innovations}).

\begin{table}[H]
\centering
\caption{Representative tactical mechanisms introduced by the LLM mutation process during evolution. None of these mechanisms existed in the seed strategies ($\times$).}
\label{tab:innovations}
\small
\begin{tabular}{@{}lp{5.0cm}c@{}}
\toprule
\textbf{Mechanism} & \textbf{Description} & \textbf{In seeds} \\
\midrule
H-DWA & Lookahead search: enumerates 23 movement headings and simulates 1.5\,s consequences & $\times$ \\
A-Lock & Locks role assignment when deep in enemy territory or during pursuit & $\times$ \\
K-Filter & EWMÍA acceleration-based interception: predicts maneuvering enemy trajectories & $\times$ \\
SQRT-VDS & Repulsion force scales with $\sqrt{\cdot}$ threat density, prevents force explosion & $\times$ \\
T-DPH & Pursuer closer to target reduces role switching (progress inertia) & $\times$ \\
A-DRL & Midline defense line adaptively retreats based on alive teammate count & $\times$ \\
\bottomrule
\end{tabular}
\end{table}

For example, H-DWA adapts the Dynamic Window Approach \cite{fox1997dwa} to discrete single-step lookahead by evaluating 23 movement headings over a 1.5-second horizon---a mechanism entirely absent from the seeds.

\section{Conclusion}

We extended FAMOU \cite{li2026famou}, a self-evolving coding-agent framework, to adversarial multi-agent games by co-evolving the evaluation process alongside the strategies themselves. Our experiments on 3v3 maritime capture-the-flag demonstrate that FAMOU consistently outperforms both OpenEvolve and ShinkaEvolve under two backbone LLMs, with the strongest variant achieving a 68.0\% win rate across ten benchmark opponents. Beyond performance gains, the evolution process generates tactical structures---such as lookahead search (H-DWA) and EWMA-based interception (K-Filter)---entirely absent from the seed strategies, providing evidence that LLMs can serve as directed mutation operators capable of producing nontrivial algorithmic innovations.

The practical impact of these results is further validated in the AAMAS 2026 MCTF Competition, where the FAMOU-evolved strategy placed \textbf{1st in the hardware round-robin} and \textbf{3rd in simulation}, demonstrating effective sim-to-real transfer.

A key insight from our work is that \emph{evaluator design may matter more than seed quality}: the same starting strategies produce substantially different outcomes depending on whether the evaluation process adapts alongside the evolving population. In adversarial settings where the opponent landscape shifts continuously, static evaluation leads to overfitting and premature convergence, while co-evolutionary evaluation maintains selective pressure toward general robustness.

\textbf{Limitations.}
Our experiments are conducted on a single game domain (MCTF), and ablation experiments use single runs; multi-run ablations and additional game domains are needed for stronger conclusions. Game simulation dominates runtime (each full benchmark checkpoint requires ${\sim}$10 hours), and over 30,000 LLM calls represent a non-negligible computational cost.

\textbf{Future directions.}
Extending FAMOU to other adversarial tasks (e.g., RoboCup, StarCraft micromanagement) and incorporating multi-objective Pareto-frontier evaluation are important next steps. More broadly, these results suggest that the co-evolution of evaluation criteria and solution candidates---a principle well-established in evolutionary computation---deserves renewed attention in the era of LLM-driven program synthesis.

\FloatBarrier
\bibliography{references}

\newpage
\appendix
\setcounter{secnumdepth}{2}

\begin{center}
\large\textbf{Appendix: Supplementary Material}
\end{center}
\vspace{0.5em}

\section{MCTF Task Specification}
\label{app:task}

\subsection{Game Environment}
MCTF 2026 is a 3v3 maritime capture-the-flag game played on a 160\,m $\times$ 80\,m rectangular field. Table~\ref{tab:app_env} summarizes the environment parameters.

\begin{table}[!ht]
\centering
\caption{MCTF 2026 environment specification.}
\label{tab:app_env}
\small
\begin{tabular}{@{}lp{5.0cm}@{}}
\toprule
\textbf{Parameter} & \textbf{Value} \\
\midrule
Field size & 160\,m $\times$ 80\,m \\
Territory boundary & Midline at $x = 80$ \\
Teams & 3 blue vs.\ 3 red agents \\
Blue flag home & $(0, 40)$ \\
Red flag home & $(160, 40)$ \\
Scoring corners (blue) & $(0, 0)$ and $(0, 80)$, 20\,m radius \\
Flag grab range & 10\,m from enemy flag ($+0.1$ pts) \\
Capture reward & Carry flag to own corner ($+1.0$ pts) \\
Tag range & 10\,m in own territory (60\,s cooldown) \\
PowerPlay & ${\sim}$20\% of time, one random enemy disabled \\
Match duration & 600 seconds \\
Max score per match & 20 captures \\
Out-of-bounds penalty & Auto-tag + 50\% speed reduction \\
\bottomrule
\end{tabular}
\end{table}

\subsection{Agent Interface}
Each strategy implements an \texttt{Agent\_0} class inheriting from \texttt{BaseAgentPolicy}, with a single decision method:

\begin{lstlisting}[style=python]
class Agent_0(BaseAgentPolicy):
    def compute_action(self, obs, info) -> int:
        # obs: observation vector
        # info: dict with global_state
        # Returns: action 0-23
\end{lstlisting}

\subsection{Observation Space}
The global state dictionary (\texttt{info[self.id]["global\_state"]}) provides the following keys:

\begin{table}[!ht]
\centering
\caption{Observation space: per-agent and global keys.}
\label{tab:app_obs}
\small
\begin{tabular}{@{}llp{3.2cm}@{}}
\toprule
\textbf{Key} & \textbf{Type} & \textbf{Description} \\
\midrule
\multicolumn{3}{@{}l}{\emph{Per-agent keys} (indexed by \texttt{(agent\_id, key)}):} \\
\texttt{pos} & $[x, y]$ & Position coordinates \\
\texttt{heading} & float & Current heading (degrees) \\
\texttt{speed} & float & Current speed \\
\texttt{has\_flag} & bool & Carrying enemy flag \\
\texttt{on\_side} & bool & In own territory \\
\texttt{is\_tagged} & bool & Currently tagged (respawning) \\
\texttt{is\_disabled} & bool & Disabled by PowerPlay \\
\texttt{tagging\_cooldown} & float & Seconds until can tag again \\
\midrule
\multicolumn{3}{@{}l}{\emph{Global keys}:} \\
\texttt{blue\_flag\_pos} & $[x, y]$ & Blue flag position \\
\texttt{red\_flag\_pos} & $[x, y]$ & Red flag position \\
\texttt{blue\_team\_score} & float & Blue team total score \\
\texttt{red\_team\_score} & float & Red team total score \\
\bottomrule
\end{tabular}
\end{table}

\subsection{Action Space}
The action space consists of 24 discrete actions (Table~\ref{tab:app_actions}). Actions 0--22 move at speed 1.0 with the corresponding heading offset; action 23 stops the agent.

\begin{table}[!ht]
\centering
\caption{Discrete action space: 23 heading actions + stop.}
\label{tab:app_actions}
\small
\begin{tabular}{@{}cl@{}}
\toprule
\textbf{Action} & \textbf{Heading offset (degrees)} \\
\midrule
0--22 & $[-180, -120, -90, -60, -30, -15, -10,$ \\
      & $-5, -3, -2, -1, 0, 1, 2, 3, 5, 10,$ \\
      & $15, 30, 60, 90, 120, 180]$ \\
23    & Stop (speed = 0) \\
\bottomrule
\end{tabular}
\end{table}

The following navigation helpers are provided to all strategies:

\begin{lstlisting}[style=python]
_HEADINGS = [-180,-120,-90,-60,-30,-15,-10,-5,
             -3,-2,-1,0,1,2,3,5,10,15,30,
             60,90,120,180]

def bearing_to_action(rel_bearing, speed=1.0):
    if speed == 0: return 23
    rb = rel_bearing
    best_idx, best_diff = 0, float('inf')
    for i, h in enumerate(_HEADINGS):
        diff = abs((rb - h + 540.0) % 360.0 - 180.0)
        if diff < best_diff:
            best_diff, best_idx = diff, i
    return best_idx

def navigate_to(target, my_pos, my_heading):
    d = np.asarray(target) - np.asarray(my_pos)
    ab = math.degrees(math.atan2(d[0], d[1]))
    return bearing_to_action(
        (ab - my_heading + 180) % 360 - 180)
\end{lstlisting}

\section{Evolution Configuration}
\label{app:config}

\subsection{FAMOU Hyperparameters}
Table~\ref{tab:app_hyper} lists the hyperparameters used for all FAMOU experiments.

\begin{table}[!ht]
\centering
\caption{FAMOU hyperparameter configuration.}
\label{tab:app_hyper}
\small
\resizebox{\columnwidth}{!}{%
\begin{tabular}{@{}llll@{}}
\toprule
\textbf{Parameter} & \textbf{Value} & \textbf{Parameter} & \textbf{Value} \\
\midrule
\multicolumn{4}{@{}l}{\emph{Evolution}} \\
Max iterations & 400 & Checkpoint interval & 40 \\
Strategy & adaptive\_cluster & Seed & 42 \\
\midrule
\multicolumn{4}{@{}l}{\emph{Island model}} \\
Num islands & 4 & Island size & 50 \\
Migration topology & ring & Migration interval & 15 \\
Migration size & 2 & Reset interval & 60 \\
\midrule
\multicolumn{4}{@{}l}{\emph{Co-evolution (epoch)}} \\
Epoch interval & 50 & Deep eval top-$k$ & 3 \\
Deep eval games & 20/opp & Weakness pressure & true \\
\midrule
\multicolumn{4}{@{}l}{\emph{LLM configuration}} \\
Backbone (config 1) & Gemini-2.5-Flash & Backbone (config 2) & DeepSeek-V4-Flash \\
Temperature & 0.8 & Max tokens & 64,000 \\
Timeout & 300\,s & Mutation types & full, cross \\
\midrule
\multicolumn{4}{@{}l}{\emph{Fast evaluation}} \\
Games per opponent & 3 & Match duration & 600\,s \\
\bottomrule
\end{tabular}%
}
\end{table}

\subsection{Benchmark Opponent Pool}
Table~\ref{tab:app_opponents} lists the 10 fixed benchmark opponents used for checkpoint evaluation throughout all experiments.

\begin{table}[!ht]
\centering
\caption{Benchmark opponent pool (fixed across all experiments).}
\label{tab:app_opponents}
\small
\begin{tabular}{@{}llcp{3.0cm}@{}}
\toprule
\textbf{ID} & \textbf{Type} & \textbf{Approx.\ strength} & \textbf{Description} \\
\midrule
\multicolumn{4}{@{}l}{\emph{Seen during evolution (training opponents):}} \\
hard & Built-in & --- & Platform default hard bot \\
Balanced & Heuristic & ${\sim}100\%$ & Role assignment + potential field \\
Agile & Heuristic & ${\sim}95\%$ & Distributed auction + vortex \\
Structured & Heuristic & ${\sim}90\%$ & Three-lane + zone defense \\
Safe & Heuristic & ${\sim}85\%$ & Beam search return \\
\midrule
\multicolumn{4}{@{}l}{\emph{Unseen (held-out for evaluation only):}} \\
Reactive & Heuristic & ${\sim}0.53$ & Auction + heat-driven urgency \\
Orbital & Heuristic & ${\sim}0.73$ & Tangential sliding + density lanes \\
Vortex & Heuristic & ${\sim}0.80$ & Adaptive vortex + tactical handoff \\
Forcefield & Heuristic & ${\sim}0.92$ & Quadratic intercept + social force \\
Elite & Heuristic & ${\sim}0.96$ & P-Flow + multi-factor adaptive \\
\bottomrule
\end{tabular}
\end{table}

\section{LLM Mutation Prompts}
\label{app:prompts}

This section presents the prompt templates used to instruct the backbone LLM during evolution.

\subsection{Task Description Prompt}
The following task description is included in every mutation prompt, providing the LLM with game rules and interface specification:

\begin{lstlisting}[style=python,basicstyle=\fontsize{6}{7.2}\selectfont\ttfamily]
OBJECTIVE:
Design a heuristic policy that MAXIMIZES AVERAGE
CAPTURES PER GAME in MCTF2026.
*** RANKING is by TOTAL CAPTURES -- not win/loss! ***
A 5-0 win is 5x more valuable than 1-0 win.

KEY INSIGHT -- SCORING EFFICIENCY:
1. MINIMIZE WASTED TIME:
   - Shortest path to enemy flag and back
   - After scoring, immediately cycle next attacker
   - Respawning wastes ~15s; avoid unnecessary tags
2. SMART ROLE ALLOCATION:
   - 2 attackers + 1 defender is a solid baseline
   - During POWERPLAY (3v2): shift to 3 attackers
3. EFFICIENT FLAG RUNS:
   - Go to NEAREST scoring corner (0,0) or (0,80)
   - Use lane switching to avoid defenders
4. COORDINATION OVER RAW SPEED:
   - One agent drawing defenders while another grabs

GAME RULES:
FIELD: 160m x 80m. Blue x<80, Red x>80.
Blue Flag: (0,40) | Red Flag: (160,40)
Scoring Zones: 20m radius around (0,0) and (0,80).
TAG: 10m in own territory -> respawn (60s cooldown)
FLAG GRAB: within 10m of enemy flag (+0.1 pts)
CAPTURE: flag to own corner (+1.0 pts)
POWERPLAY: ~20% time, one enemy disabled

INTERFACE:
Class: Agent_0 (inherits BaseAgentPolicy)
Method: compute_action(obs, info) -> int (0-23)
Actions 0-22: Speed 1.0, heading offsets
Action 23: Stop
HARD REQUIREMENTS: pure Python + numpy only.
\end{lstlisting}

\subsection{Full Rewrite Mutation Prompt}
For structural changes, the LLM is asked to produce a complete rewritten strategy file. We use five prompt variants, sampled stochastically:

\begin{enumerate}
\item \textbf{Default}: ``Rewrite the program to improve its performance on the specified metrics.''
\item \textbf{Different algorithm}: ``Design a completely different algorithm approach to solve the same problem.''
\item \textbf{Context-motivated}: ``Create a novel algorithm that draws inspiration from the provided context programs but implements a fundamentally different approach.''
\item \textbf{Structural redesign}: ``Redesign the program with a focus on restructuring the core algorithmic components.''
\item \textbf{Parametric}: ``Focus on tuning constants, thresholds, and parameters while keeping the overall algorithmic structure.''
\end{enumerate}

All variants require the LLM to preserve the \texttt{EVOLVE-BLOCK-START}/\texttt{END} markers and maintain the same input/output interface.

\section{Seed Strategy Code}
\label{app:seed}

Listing~\ref{lst:seed} shows the complete seed strategy used to initialize all experiments. This simple heuristic implements a 2-attacker + 1-defender scheme with no evasion, no coordination, and no advanced tactics---intentionally providing room for the evolutionary algorithm to improve.

\begin{lstlisting}[style=python,caption={Complete seed strategy (\texttt{seed\_simple.py}, 217 lines). Role assignment is static (first agent = defender), attackers navigate directly to the enemy flag, and the defender patrols near its own flag.},label={lst:seed},basicstyle=\fontsize{5.5}{6.6}\selectfont\ttfamily]
import math
import numpy as np
from pyquaticus.base_policies.base_policy import BaseAgentPolicy

# EVOLVE-BLOCK-START

_HEADINGS = [-180,-120,-90,-60,-30,-15,-10,-5,
             -3,-2,-1,0,1,2,3,5,10,15,30,
             60,90,120,180]
BLUE_CORNERS = [np.array([0.,0.]), np.array([0.,80.])]
BLUE_FLAG_HOME = np.array([0.0, 40.0])
RED_FLAG_HOME = np.array([160.0, 40.0])
FIELD_W, FIELD_H = 160.0, 80.0
TAG_RANGE = 10.0

def angle180(a):
    a = a % 360.0
    return a - 360.0 if a > 180.0 else a

def bearing_to_action(rel_bearing):
    rel_bearing = angle180(rel_bearing)
    best_idx, best_diff = 0, float('inf')
    for i, h in enumerate(_HEADINGS):
        d = abs(angle180(rel_bearing - h))
        if d < best_diff:
            best_diff, best_idx = d, i
    return best_idx

def nav_to(target, my_pos, my_heading):
    d = np.asarray(target) - np.asarray(my_pos)
    if np.linalg.norm(d) < 0.5:
        return 23  # arrived
    abs_bear = math.degrees(
        math.atan2(d[0], d[1])) % 360.0
    rel_bear = angle180(abs_bear - my_heading)
    return bearing_to_action(rel_bear)

def nearest_corner(pos, corners):
    return min(corners,
        key=lambda c: np.linalg.norm(c - np.asarray(pos)))

class Agent_0(BaseAgentPolicy):
    def __init__(self, id, env):
        super().__init__(id, env)
        self.team_ids = []
        self.enemy_ids = []
        self.is_blue = True
        self.my_corners = BLUE_CORNERS
        self.enemy_flag_home = RED_FLAG_HOME
        self.my_flag_home = BLUE_FLAG_HOME
        self.role = "attacker"
        self._initialized = False

    def _init_once(self, gs):
        if self._initialized: return
        idx = int(self.id.split('_')[1])
        if idx < 3:
            self.is_blue = True
            self.team_ids = [f'agent_{i}' for i in range(3)]
            self.enemy_ids = [f'agent_{i}' for i in range(3,6)]
        else:
            self.is_blue = False
            self.team_ids = [f'agent_{i}' for i in range(3,6)]
            self.enemy_ids = [f'agent_{i}' for i in range(3)]
        # First agent = defender, others = attacker
        self.role = "defender" if self.id == self.team_ids[0] \
                    else "attacker"
        self._initialized = True

    def compute_action(self, obs, info):
        gs = info.get(self.id, {}).get('global_state', {})
        if not gs: return 23
        self._init_once(gs)
        my_pos = np.asarray(gs.get((self.id,'pos')), dtype=float)
        my_heading = float(gs.get((self.id,'heading'), 0.0))
        my_has_flag = bool(gs.get((self.id,'has_flag'), False))
        if gs.get((self.id,'is_tagged'), False): return 23

        # If carrying flag -> nearest scoring corner
        if my_has_flag:
            return nav_to(nearest_corner(my_pos, self.my_corners),
                          my_pos, my_heading)
        # Chase enemy flag carrier on own side
        for eid in self.enemy_ids:
            if gs.get((eid,'has_flag'), False):
                epos = gs.get((eid,'pos'))
                if epos is not None and \
                   bool(gs.get((self.id,'on_side'), True)):
                    return nav_to(np.asarray(epos), my_pos, my_heading)
        # Powerplay -> all attack
        enemy_active = sum(1 for eid in self.enemy_ids
            if not gs.get((eid,'is_tagged'),False)
            and not gs.get((eid,'is_disabled'),False))
        if enemy_active < 3:
            flag = gs.get('red_flag_pos' if self.is_blue
                          else 'blue_flag_pos')
            return nav_to(np.asarray(flag) if flag
                          else self.enemy_flag_home,
                          my_pos, my_heading)
        # Role-based behavior
        if self.role == "defender":
            return self._defend(gs, my_pos, my_heading)
        return self._attack(gs, my_pos, my_heading)

    def _attack(self, gs, my_pos, my_heading):
        flag = gs.get('red_flag_pos' if self.is_blue
                      else 'blue_flag_pos')
        return nav_to(np.asarray(flag) if flag
                      else self.enemy_flag_home,
                      my_pos, my_heading)

    def _defend(self, gs, my_pos, my_heading):
        # Chase closest enemy intruder on our side
        closest, closest_d = None, float('inf')
        for eid in self.enemy_ids:
            if gs.get((eid,'is_tagged'),False): continue
            epos = gs.get((eid,'pos'))
            if epos is None: continue
            epos = np.asarray(epos, dtype=float)
            on_our = (epos[0]<80) if self.is_blue else (epos[0]>80)
            if on_our:
                d = np.linalg.norm(epos - my_pos)
                if d < closest_d:
                    closest_d, closest = d, epos
        if closest is not None and closest_d < 50.0:
            return nav_to(closest, my_pos, my_heading)
        # Patrol near own flag
        patrol = np.asarray(self.my_flag_home)
        offset = np.array([15.,0.]) if self.is_blue \
                 else np.array([-15.,0.])
        patrol = np.clip(patrol + offset, [2,2], [158,78])
        if np.linalg.norm(patrol - my_pos) < 3.0:
            return 23
        return nav_to(patrol, my_pos, my_heading)

# EVOLVE-BLOCK-END
\end{lstlisting}

\section{Evolved Champion Strategy (Excerpt)}
\label{app:champion}

Listing~\ref{lst:champion} shows an excerpt from a late-stage evolved champion (epoch~8, iteration~258), highlighting three LLM-generated innovations absent from the seed:

\begin{enumerate}
\item \textbf{Dynamic role assignment} (\texttt{\_get\_dynamic\_roles}): Roles are reassigned every tick based on agent proximity to strategic points, replacing the seed's static first-agent-is-defender scheme.
\item \textbf{Avoidance waypoint calculation} (\texttt{calculate\_avoidance\_waypoint}): Flag carriers and primary attackers compute temporary waypoints to steer around enemies blocking their path, a mechanism entirely absent from the seed.
\item \textbf{Defender patrol system}: The defender cycles through multiple patrol points along the midline (configurable via \texttt{DEFENDER\_PATROL\_Y\_LANES}), rather than camping at a single fixed position.
\end{enumerate}

The mutation log (preserved as comments) documents the LLM's reasoning:

\begin{lstlisting}[style=python,caption={Evolved champion excerpt (epoch 8, iteration 258). Comments at the top document the LLM's mutation reasoning. Only key evolved components are shown; the full file is ${\sim}$500 lines.},label={lst:champion},basicstyle=\fontsize{5.5}{6.6}\selectfont\ttfamily]
# Mutation Level: STRUCTURAL
# Specific Changes:
#   a. Reintroduce Dynamic Role Assignment:
#      Replaced static role assignment with dynamic
#      system. Roles ('defender', 'primary_attacker',
#      'support_attacker') assigned each tick based on
#      agent proximity to strategic points.
#   b. Refine Attacker Coordination:
#      Primary attacker focuses on flag with evasion.
#      Support attacker trails or positions for relay.
#   c. Defender Patrol: Cycles through predefined
#      patrol points along midline.
#   d. Robust Avoidance: Waypoint-based steering
#      around detected enemies "in the way".

# --- Evolved Constants (tuned by LLM) ---
DEFENDER_PATROL_X_OFFSET = 10.0
DEFENDER_PATROL_Y_LANES = [20.0, 40.0, 60.0]
SUPPORT_ATTACKER_TRAIL_DIST = 30.0
AVOID_RANGE = 30.0
IN_THE_WAY_ANGLE_THRESHOLD = 30.0
AVOID_STEER_ANGLE = 45.0
AVOID_WAYPOINT_DIST = 50.0

def calculate_avoidance_waypoint(
        my_pos, primary_target, enemy_pos):
    """Temporary waypoint to steer around enemy."""
    vec_to_enemy = enemy_pos - my_pos
    vec_to_target = primary_target - my_pos
    if np.linalg.norm(vec_to_target) < 1.0:
        return primary_target
    bear_enemy = math.degrees(
        math.atan2(vec_to_enemy[0], vec_to_enemy[1]))
    bear_target = math.degrees(
        math.atan2(vec_to_target[0], vec_to_target[1]))
    angle_diff = angle180(bear_enemy - bear_target)
    # Steer opposite side from enemy
    if angle_diff > 0:
        new_bear = bear_target + AVOID_STEER_ANGLE
    else:
        new_bear = bear_target - AVOID_STEER_ANGLE
    wp_x = my_pos[0] + AVOID_WAYPOINT_DIST * \
           math.sin(math.radians(new_bear))
    wp_y = my_pos[1] + AVOID_WAYPOINT_DIST * \
           math.cos(math.radians(new_bear))
    return np.clip([wp_x, wp_y], [2,2], [158,78])

class Agent_0(BaseAgentPolicy):
    def _get_dynamic_roles(self, gs):
        """Assign roles based on current positions."""
        active = []
        for tid in self.team_ids:
            if not gs.get((tid,'is_tagged'), False) \
               and not gs.get((tid,'is_disabled'), False) \
               and not gs.get((tid,'has_flag'), False):
                pos = gs.get((tid, 'pos'))
                if pos is not None:
                    active.append(
                        (tid, np.asarray(pos, dtype=float)))
        roles = {}
        if not active: return roles
        # Defender: closest to own flag
        active.sort(key=lambda x:
            np.linalg.norm(x[1] - self.my_flag_home))
        roles[active[0][0]] = "defender"
        # Primary attacker: closest to enemy flag
        remaining = active[1:]
        if remaining:
            remaining.sort(key=lambda x:
                np.linalg.norm(x[1] - self.enemy_flag_home))
            roles[remaining[0][0]] = "primary_attacker"
        if len(remaining) > 1:
            roles[remaining[1][0]] = "support_attacker"
        return roles
\end{lstlisting}

\end{document}